\documentclass[letterpaper]{article} 
\usepackage{aaai24}  
\usepackage{times}  
\usepackage{helvet}  
\usepackage{courier}  
\usepackage[hyphens]{url}  
\usepackage{graphicx} 
\usepackage{natbib}  
\usepackage{caption} 
\usepackage{algorithm}
\usepackage{algorithmic}
\usepackage[export]{adjustbox}
\usepackage{tikz}
\usetikzlibrary{shapes,arrows}
\usepackage{newfloat}
\usepackage{multirow}
\usepackage{hhline}
\usepackage{todonotes}
\usepackage[most]{tcolorbox}
\usepackage{listings}
\usepackage{arydshln}
\RequirePackage{amsmath,amssymb,bm,amsthm}
\RequirePackage[inline, shortlabels]{enumitem}
\DeclareCaptionStyle{ruled}{labelfont=normalfont,labelsep=colon,strut=off} 
\floatstyle{ruled}
\newfloat{listing}{tb}{lst}{}
\floatname{listing}{Listing}

\def\modelname{{\scshape SyReLM}}
\def\ztitle{Frugal LMs Trained to Invoke Symbolic Solvers Achieve
Parameter-Efficient Arithmetic Reasoning}
\title{\ztitle}
\author{
    Subhabrata Dutta\equalcontrib\textsuperscript{\rm 1}, Joykirat Sing\equalcontrib\textsuperscript{\rm 2}, Ishan Pandey\equalcontrib\textsuperscript{\rm 2},\\ Sunny Manchanda\textsuperscript{\rm 3}, Soumen Chakrabarti\textsuperscript{\rm 4}, Tanmoy Chakraborty\textsuperscript{\rm 1}
}
\affiliations{
    \textsuperscript{\rm 1}IIT Delhi, India,
    \textsuperscript{\rm 2}IIIT Delhi, India,
    \textsuperscript{\rm 3}DYSL-AI, India,
    \textsuperscript{\rm 4}IIT Bombay, India\\
    subha0009@gmail.com, joykirat19166@iiitd.ac.in, ishan20304@iiitd.ac.in, manchanda.sunny@gmail.com, soumen@cse.iitb.ac.in, tanchak@ee.iitd.ac.in
}

\usepackage{bibentry}

\begin{document}

\maketitle

\begin{abstract}
Large Language Models (LLM) exhibit zero-shot mathematical reasoning capacity as a behavior emergent with scale, commonly manifesting as chain-of-thoughts (CoT) reasoning. However, multiple empirical findings suggest that this prowess is exclusive to LLMs with exorbitant sizes (beyond 50 billion parameters). Meanwhile, educational neuroscientists suggest that symbolic algebraic manipulation be introduced around the same time as arithmetic word problems to modularize language-to-formulation, symbolic manipulation of the formulation, and endgame arithmetic.
In this paper, we start with the hypothesis that much smaller LMs, which are weak at multi-step reasoning, can achieve reasonable arithmetic reasoning if arithmetic word problems are posed as a formalize-then-solve task.
In our architecture, which we call \modelname, the LM serves the role of a translator to map natural language arithmetic questions into a formal language (FL) description. A symbolic solver then evaluates the FL expression to obtain the answer.
A small frozen LM, equipped with an efficient low-rank adapter, is capable of generating FL expressions that incorporate natural language descriptions of the arithmetic problem (e.g., variable names and their purposes, formal expressions combining variables, etc.).
We adopt policy-gradient reinforcement learning to train the adapted LM, informed by the non-differentiable symbolic solver. This marks a sharp departure from the recent development in tool-augmented LLMs, in which the external tools (e.g., calculator, Web search, etc.) are essentially detached from the learning phase of the LM. \modelname{} shows massive improvements (e.g., $+30.65$ absolute point improvement in accuracy on the SVAMP dataset using GPT-J 6B model)  over base LMs, while keeping our testbed easy to diagnose, interpret and within reach of most researchers.
\end{abstract}

\section{Introduction}
\label{sec:intro}

Large Language Models (LLMs) trained on giant corpora of text, code, reasoning chains, and dialogues have recently taken centerstage (ChatGPT\footnote{\url{https://chat.openai.com/}}, BARD\footnote{\url{https://bard.google.com/}}) in not only most NLP tasks but also arithmetic and logical reasoning~\citep{GPT3, kojima2022LLMzeroShot}.  Despite major recent strides, LLMs, on their own, are resource-intensive, inefficient and unreliable devices for many rigorous tasks such as arithmetic, logic, calculus, or geometry.  Indeed, one should not expect word sequence inputs and outputs to capture such domains and tasks.  Moreover, folding such `skills' into an opaque, massive LM resists diagnosis and interpretability. Therefore, LLMs are also being trained to invoke \textit{tools} that can perform such tasks as subroutines \citep{paranjape2023art, Toolformer, Wolfram2023ChatGptWolfram}, with the LLM acting as a `glue' between user utterances and these tools.

Here, we focus on the interplay between symbolic and arithmetic reasoning tasks mediated by an LM.  Educational neuroscientists have long detected a stronger correlation between dysfunctional word problem-solving capability and symbolic algebraic reasoning than between the former and arithmetic calculation abilities \citep{NathanKY1992AlgebraWordProblem, Powell+2014AlgeraArithmetic}.  In light of fMRI evidence, this may not be surprising that different areas of the brain \citep[\S 2.2.1, \S 2.2.2]{mahowald2023dissociating} are responsible for linguistic, analytical and logical processing.  A common pedagogic trick employed by grade school teachers is to repeat arithmetic word problems with diverse numerals to see if a pupil understands the problem at the level of symbolic reasoning, as against numeric shortcuts.  Beyond grade school, the correctness of SQL queries is checked by weeding out incorrect \textit{mutant} queries using adversarially engineered database instances \citep{Chandra+2015xdata}.

\begin{figure*}[t]
\centering
\begin{tabular}{cp{.3\textwidth}}
\includegraphics[width=0.67\textwidth,valign=t]{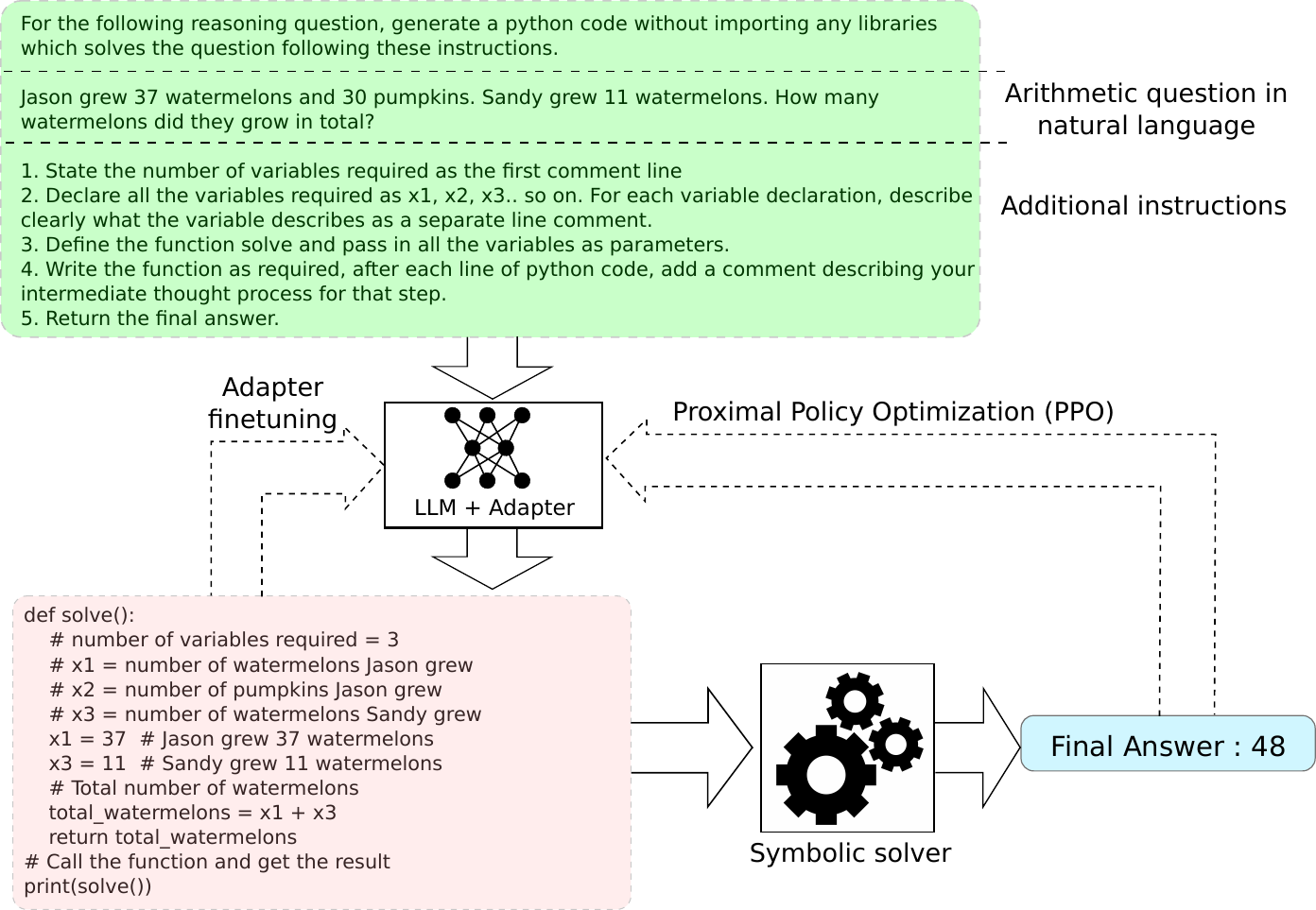}
&
\caption{{\bfseries \modelname{} architecture through an example}. Given an arithmetic word problem in natural language, we first introduce a set of instructions to generate a formal language expression (Python, in this example).
Next, a symbolic solver (a Python interpreter, in this example) executes the formal language expression to obtain the final answer. In the first stage of optimization, the adapter module is finetuned using a language modeling objective. Subsequently, we use Proximal Policy Optimization (PPO) to finetune the adapter, where the final answer and the coverage of variables needed are used as rewards. The base LM is kept frozen throughout.}
\label{fig:model-scheme}
\end{tabular}
\end{figure*}

The above discussion naturally suggests that an LM that can call an arithmetic calculator tool to solve word problems can perform better (perhaps even with a smaller model size) if it also learns to translate the word problem to a logical form in a \textit{formal language} (FL) and a {\it symbolic solver} to manipulate the FL specification. After all, the recipe often given to school children to solve word problems goes roughly like this (paraphrased for the current audience):

\begin{enumerate}[noitemsep]
\item Read the word problem to allocate variables to quantities mentioned therein.
\item Parse the text to bind some subset of variables to constants grounded in the problem statement.
\item Further parse the text to extract arithmetic relationships and constraints between variables.
\item Identify the target variable(s) whose ground value(s) will answer the question.
\item Invoke a symbolic solver to express the target variable(s) in terms of grounded variables.
\item Invoke an arithmetic calculator/interpreter to obtain values for target unknown variables.
\end{enumerate}
Figure~\ref{fig:model-scheme} shows an extended example of the above steps in action.
Dysfunction in solving word problems (such as sometimes evident in even powerful LLMs, as well as grade-school students) is often traced to a failure in one or more of the above steps.
This naturally suggests that we attempt to apply grade-school pedagogy tricks to tool-using LMs.

\paragraph{Our Contributions.}
Early LM-based word problem solvers \citep{CoT, PaLM} employed LM to manage {\it all} the above steps; however, recent LLM-based systems \citep{paranjape2023art, Toolformer} are taught to invoke tools.
Instead of burdening a gigantic LLM with language understanding, symbolic processing, and proper invocation of an arithmetic calculator tool, we propose \modelname, a system where the LM serves the role of a translator from natural language arithmetic questions into a formal language description, and then invokes a symbolic solver to evaluate the formal language expression to obtain the answer.
In contrast to gigantic LLMs that are beyond many researchers' capability to train or finetune, we use a frozen LM of modest size with a low-rank adapter~\citep{LoRA}.
We adopt policy-gradient reinforcement learning to train the adapted LM, along with the non-differentiable symbolic solver. This marks a sharp departure from the recent development in tool-augmented LLMs, in which the external tools are essentially detached from the learning phase of the LM.  \modelname{} shows strong improvements over base LMs, while keeping our testbed easy to diagnose, interpret and within the reach of most researchers.  While still far short of LLMs trained with huge corpora and human feedback \citep{openai2023gpt4, zheng2023progressivehint}, \modelname{} demonstrates strong performance boosts and establishes the validity of middle-school math pedagogy for AI pupils. On the SVAMP arithmetic reasoning dataset, \modelname\ improves upon the base $k$-shot accuracy of GPT-J 6 billion \citep{gpt-j} (Vicuna 13 billion\footnote{\url{https://lmsys.org/blog/2023-03-30-vicuna/}}) by \(31.6\ (27.12)\) absolute points, outshining recent methods like Toolformer \citep{Toolformer} and ART \citep{paranjape2023art} by large margins. On SVAMP and GSM8K datasets, \modelname\ makes Vicuna 13B perform comparable to GPT-3.5 (standard 8-shot prompting). We release the code and data for reproducibility~\footnote{\url{https://github.com/joykirat18/SYRELM}}.

\section{Related Work}
\label{sec:Rel}

Given their original purpose of modelling word correlations, early LMs like BERT \citep{devlin-etal-2019-bert} were found astonishingly capable of arithmetic but within restricted ranges \citep{wallace2019lmnumeracy}.  They found that GloVe and word2vec encode numbers up to about $\pm1000$ with reasonable accuracy and that character-level embeddings as in ELMo \citep{Peters+2018ELMO} are (understandably) more precise; however, BERT, which uses sub-word units, is less exact.  Around the same time, \citet{trask2019NALU} explored the design of ``neural arithmetic logic units'' (NALUs) that specialized in interconverting between lexicalized quantities and linear activations, with the goal of manipulating quantities beyond those encountered in a textual corpus.  Middle-school teachers expect students to solve word problems no matter whether John gave Jane 12 pencils or 13,625,018.  However, LLMs like T5 and GPT-3 were getting so good at covering common benchmarks like GSM8K \cite{cobbe2021gsm8k} --- and even performing addition \citep{zhou2022InContextAlgorithm} and multiplication \citep{Narayanan2022ChatGptMultiply} `properly' --- that it took a few months before arithmetic calculator tools were harnessed to LLMs. Prominent early systems to do so are Toolformer \citep{Toolformer}, ART \citep{paranjape2023art}, Program-of-Thoughts Prompting \cite{PoT} and Program-aided Language Models \cite{PAL}, which take somewhat different paths toward similar goals.

Toolformer uses GPT-J \cite{mesh-transformer-jax} as the base LM with 6.7 billion parameters, which is amenable to fine-tuning using modest hardware.  It outperforms the much larger GPT-3 model at various tasks using the following tools: a calculator, a calendar,  a question-answering (QA) tool, a traditional BM25-based
Wikipedia search tool and a machine translation tool.  For our purposes, the only relevant tool is the calculator, and the only relevant datasets are SVAMP \cite{patel-etal-2021-nlp}, MAWPS \footnote{\url{https://github.com/sroy9/mawps}} and ASDiv \cite{miao-etal-2020-diverse}.  We will compare \modelname{} against some of these tasks (\S \ref{sec:results}).  Toolformer does not use a symbolic solver in conjunction with a numeric calculator like \modelname.  ART  uses GPT-3 with 175 billion parameters, which means it already has access to substantial reasoning ability packaged into the massive but opaque LM.

\textcolor{black}{Program-of-Thoughts Prompting \cite{PoT} and Program-aided Language Models \cite{PAL} use LLMs as symbolic solvers to read natural language problems and generate programs as intermediate steps, offloading computation steps to a Python interpreter. 
PoT and PAL depend heavily on prompt engineering with fixed LLMs and not efficient LM adaptation.
\citet{qiao_making_2023} used RLHF \cite{2304.05302} framework to teach LLMs how to selectively use tools. First, the model is taught how to invoke tools, and then reinforcement learning is used in selective tool usage.}
 
The latest development in this direction has been the integration of Wolfram Alpha `plugins' into ChatGPT \citep{Wolfram2023ChatGptWolfram}. This collaboration incorporated diverse tools from various domains that led to prompting ChatGPT to compose queries suited to Wolfram Alpha, collect the latter's output, and then either relay or transcribe or compose it with further operations.  While Wolfram+ChatGPT wide-ranging tools will have a great impact, our focus here is on the specific synergy between symbolic interpretation and arithmetic problem-solving skills.

\citet{zheng2023progressivehint} obtained SOTA performance by starting with GPT-4 \citep{openai2023gpt4}, it iteratively collects putative answer in a set $S$ and continue with rounds of extra prompt ``(Hint: The answer is near to $S$)'' until the LLM's answer stabilizes. The gains might be from a Monty Hall effect \citep{wiki:Monty_Hall_problem}, but this method clearly moves away from pedagogic wisdom \citep{Gunzelmann2005MultipleChoice}, creating an approximation to a multiple choice test when none existed in the first place.  Our approach aims to move in a direction where behaviour modification is directly interpreted and justified.


\section{Arithmetic Reasoning: Natural to Formal Language}
\label{sec:formal-language-expression}

As we discussed earlier, expressing the reasoning steps required to solve a given arithmetic problem in some FL (parsable pseudocode, python, lambda calculus, etc.) shifts the burden of deterministic evaluation from the LM itself to some dedicated solver. However, the choice of FL is tricky due to multiple factors. First, the underlying LM should be exposed to generate expressions in the chosen FL in its pretraining stage. For example, {Codex}\footnote{\url{https://platform.openai.com/docs/guides/code}} \citep{chen2021evaluating} outperforms GPT-3 \citep{GPT3} by huge margins while using Program-of-Thoughts (PoT) prompting~\citep{PoT}. Second, there is a trade-off between the complexity of the FL vs the size of the generated expression; an FL with a restricted vocabulary would require more steps to express the problem. The LM responsible for generating the FL expression might fail with longer reasoning chains. On the other hand, a large vocabulary creates a harder decision choice at each step due to increased possible options at each step.

Previous studies have shown the effectiveness of Python as a choice of FL for solving arithmetic problems with LMs~\citep{PAL,PoT}. In designing \modelname, we follow the same strategy with instruction-tuned models capable of generating Python codes. However, we adopt a few additional instructions to facilitate reasoning-specific rewards in downstream policy-gradient optimization. As shown in Figure~\ref{fig:model-scheme}, we instruct the LM in \modelname{} to list all the necessary variables, along with their contextual meaning in natural language comments. We also add a CoT-like flavor by asking the model to generate interpretive natural language comments explaining the reasoning behind the Python statements. This aligns the two modes of reasoning: CoT vs. PoT. While pure CoT provides step-by-step and explainable reasoning to solve a problem, one cannot separate the computation from language generation. On the other hand, PoT provides a detachable computation graph at the expense of reduced reasoning explanation. By blending these two, we enforce that the LM actually generalizes across two different modes of reasoning. In the later section, we describe the usage of the additional instructions (e.g., variable names, values, etc.) while discussing the policy-gradient optimization step.

\paragraph{Parsable Pseudocodes.} Not all LMs are optimized for Python code generation. Given the rich syntax of Python, finetuning such models with additional Python scripts is beyond the computational capacity accessible to most researchers. Instead, in \modelname, we formulate a simple FL prescription for such models. 

Figure~\ref{fig:pseudocode-example} shows an example of such a pseudocode. From the perspective of the language model, we instruct the LM tokenizer not to break the tokens corresponding to special symbols like {\bf [find]}, {\bf [add]}, etc., into subwords. We use a simple parser that reads these statements top to bottom, collects and stores the variables ({\bf var1}, {\bf var2}, etc.) along with their values (commented numerical values after each {\bf [find]} statement), and evaluates expressions containing arithmetic operations ({\bf [add]}, {\bf [subtract]}, etc.). 
The LM acts in synergy with the parser; after generating a statement containing an arithmetic expression (e.g., {\bf var4 = [subtract](var1,var3)} in Figure~\ref{fig:pseudocode-example}), the language model halts, and the parser is called for evaluation. The output of the parser is then appended as a comment (e.g., \(\#\ 7-10=-3\)), and the LM resumes generation from there. This {\em halt-compute-resume} regime of incorporating an external tool with the LM is somewhat similar to Toolformer~\citep{Toolformer}. However, a key distinction emerges at later stages with incorporating policy-gradient optimization.

We formally introduce the general problem as follows.{ Given a natural language arithmetic problem as a token sequence \(T_Q\) and a set of instructions as a token sequence \(T_{Inst}\), the LM maps \([T_Q, T_{Inst}]\) to an FL token sequence \(T_{FL}\)} that is then deterministically evaluated using a symbolic solver \(\bf S\).

\begin{figure}[!t]
\begin{tikzpicture}[
    greenbox/.style={draw=green!5, thick, fill=green!10, text width=\hsize-12pt, align=left, inner sep=6pt},
    redbox/.style={draw=red!5, thick, fill=red!10, text width=\hsize-12pt, align=left, inner sep=6pt},
    arrow/.style={thick, ->, >=stealth},
    line/.style={dashed, gray}
  ]
  \node[greenbox] (green) { \small
  Jerry had 7 action figures on a shelf in his
room. Later he added some more action
figures to the shelf and removed 10 of the old
ones. If there are 8 action figures on his shelf
now How many action figures did he add to
the shelf?
  };
  \node[redbox, below=3mm of green] (red) {\scriptsize \ttfamily \bfseries
var1 = [find](figures on shelf) \# 7, \\
var2 = [find](figures added) \# ?, \\
var3 = [find](figures removed) \# 10, \\
var4 = [subtract](var1, var3) \# 7 - 10 = -3, \\
var5 = [find](figures now) \# 8, \\
var6 = [subtract](var5, var4) \# 8-(-3) = 11, \\
\, [return] (var6) \# 11  };
  \draw[arrow] (green.south) -- (red.north);
\end{tikzpicture}
    \caption{Example of parsable pseudocode generation from LMs. Given the natural language arithmetic question (green), it first defines the required variables via {\bf [find]} statements, followed by the computation statements using arithmetic operations like {\bf [add]}, {\bf [subtract]}, etc. The final answer is declared via the {\bf [return]} statement.} \label{fig:pseudocode-example}
\end{figure}

\section{Training with Adapters}
\label{sec:training-method}

There exists a dramatic gap in expressiveness between models like GPT-3, and PaLM vs their smaller counterparts in the 1B-50B parameter scale, especially when the task is further away from simple language generation. Evidently, we need base LMs to \textit{adapt} to the FL generation task via some form of finetuning. However, traditional finetuning is problematic in this scenario for three major reasons --- (i) {\bf loss of generalizability}: the LM will lose its generic language modeling ability and get biased towards the small set of arithmetic problems presented, (ii) {\bf computational cost}: full finetuning would incur computational expenses beyond the capability of most practitioners, and (iii) {\bf practical deployment challenges}: deploying highly task-specific models for different tasks would be extremely cumbersome compared to having one fixed large model with small task-specific modules on top of it. To address these challenges, we make use of a Low-Rank Adapter(LoRA) \citep{LoRA}. 

As shown in Figure~\ref{fig:model-scheme}, the training procedure is two-staged. First, the adapter-augmented LM is finetuned to translate the natural language arithmetic problems to FL expressions. We further optimize the model along with the symbolic solver using Proximal Policy Optimization.

\paragraph{LoRA Finetuning with LM Objective.}
This first stage of optimization aligns the model to generate FL expressions \(T_{FL}\) from the instruction augmented arithmetic question \([T_Q, T_{Inst}]\). 
The objective can be expressed as
\begin{align}
\max_{ \theta_{LoRA}}\sum_{[T_Q, T_{Inst}], T_{FL}} \sum_{y\in T_{FL}} 
\log(P(y|[T_Q, &T_{Inst}], \notag \\[-3ex] & \theta, \theta_{LoRA}, {\bf S}))
\label{eq:lm-finetune}
\end{align}
where \(\theta\) and \(\theta_{LoRA}\) are the LM and LoRA parameters, respectively. \(\bf S\) denotes the (optional) symbolic solver that only comes into play while generating pseudocode sequences, as described in Section~\ref{sec:formal-language-expression}. 


\paragraph{Proximal Policy Optimization.}
Typically, a PPO setup contains three constituent modules: a policy model \(\pi_\theta\), a reference model \(\pi_{ref}\), a reward function \(r_t\), and a value function \(V(s_t)\) describing the reward at step \(t\) and value at state \(s_t\). \(\pi_\theta\) and \(\pi_{ref}\) define probability distributions over actions \(a_t\) given the state \(s_t\). We define an advantage estimator \(\hat{A}_t\) at each timestep \(t\) over a trajectory of timesteps \(0\) to \(T\) as follows:
\begin{equation}
    \hat{A}_t = \sum_{i=0}^{T-t+1} (\gamma\lambda)^i(r_{t+i}+\gamma V(s_{t+i+1})-V(s_{t+i}))
\end{equation}
where \(\gamma,\lambda\in(0,1]\) are hyperparameters controling the bias-variance trade-off. The PPO objective can be described as,
\begin{equation}
\small
    \max_\theta \left({E}_t\left [\frac{\pi_\theta (a_t|s_t)}{\pi_{ref} (a_t|s_t)}\hat{A}_t\right] - \beta {E}_t\left [{KL}[\pi_{ref}(\cdot|s_t), \pi_\theta(\cdot|s_t)]\right] \right )
\end{equation}
where \({KL}(\cdot,\cdot)\) signifies the KL-divergence, and $\beta$ is an adaptive hyperparameter controlling the KL-penalty. Intuitively, this penalty term drastically restricts the policy from diverging from the reference model. In our setting, we initialize \(\pi_\theta\) and \(\pi_{ref}\) with the adapter-augmented LM finetuned using LM objective presented in Eq. \ref{eq:lm-finetune}. Timesteps of trajectories are defined per token. The value function \(V(s_t)\) is initialized as an MLP head on top of the policy model. 

\paragraph{Reasoning-specific Reward for PPO.} The reward function \(r_t\) requires some novel approach to fit with our final goal of enhancing the reasoning ability of the LM. Typically, in the Reinforcement Learning from Human Feedback (RLHF) setting (where the PPO algorithm has found its most promising usage in conjunction with LMs), a separate LM is used to encode the human preferences from the feedback data. In our case, however, such an approach is not possible. On the other hand, we seek to enhance the LM's ability to use the symbolic solver faithfully. Therefore, the reward function can be straightforwardly defined using the correctness of the result generated by the symbolic solver. This, however, provides very little direct information about the quality of the generated FL. Intuitively, learning with such sparse information would require a vast amount of high-quality data and computing time. To tackle this, we incorporate rewards based on the generated FL description as follows:
\begin{enumerate}
    \item Successful compilation of generated Python program; \(R_1=R_{max}\) if successful, \(0\) otherwise.
    \item The absolute difference between the number of variables passed to the solve function (\(v_{gen}\)) in the generated program vs. in the gold program (\(v_{gold}\)), a larger difference signifies smaller reward; \(R_2=R_{max}\left(1-\frac{v_{gen}-v_{gold}}{v_{gold}}\right)\).
    \item \(R_3\): Number of matching arithmetic operators (\(+, -, *, /\)) between generated and gold program; \(+R_{max}\) for each matching operator, \(-R_{max}\) for each missing operator, and \(-0.5R_{max}\) for unnecessary operators generated.
    \item Absolute difference between the gold answer (\(y_{gold}\)) and the generated answer (\(y_{gen}\)) scaled by the gold answer; \(R_4=R_{max}\left(1-\frac{ y_{gen}-y_{gold}}{y_{gold}}\right)\).
\end{enumerate}
Here, \(R_{max}\) is the maximum possible reward. We set it to \(1\) upon experimenting. Total reward is then \(r = {R_1+R_2+R_3+R_4}\).
Such a diverse set of rewards provides a richer set of learning feedback for the policy model. One can intuitively map the reward ordering (\(R_1\) to \(R_4\)) as stages to learn to reason -- first, the policy model learns to generate compilable programs, then to identify the correct set of variables to use, then to select the correct set of arithmetic operations, and finally, to generate the full program that would lead to the correct answer. Compared to rewarding only the correctness of the final answer, our proposed reward function enables demarcating the capability of the policy model even while generating wrong answers.

\section{Experimental Setup}
\label{sec:experiments}

\paragraph{Training Dataset.} Given the novel setting of the FL expressions and instructions needed to train the adapters, we developed our own dataset. We built upon three existing mathword problem datasets: ASDiv~\citep{asdiv}, MAWPS~\citep{asdiv}, and Math23k~\citep{math23k}. Since the original Math23K dataset is in Mandarin, we used an English-translated version. To reduce manual curation effort, we employed InstructGPT~\citep{RLHF-2} to generate a noisy version of the pseudocode and Python programs to solve the selected problems. A manually curated 8-shot prompt was used with InstructGPT 175B version. Out of the initial 12,876 examples, we manually discarded instances with incorrect answers or flawed FL expressions. This process yielded 7,920 problems with pseudocode and 6,665 with Python programs, forming the training data for our models and constructed k-shot prompts for the baselines used in different experiments.
Further details of the training dataset can be found in Appendix~\ref{app:TrainingDataDetails}.

\paragraph{LMs Used.} 
In Section~\ref{sec:formal-language-expression}, we mentioned that different LMs possess different capacities for generating FL expressions given their pretraining data/objectives. We employ two LM classes for experimentation: Vicuna 13B (instruction-tuned version of the Llama 13B model~\citep{llama}) to generate Python, and GPT-J 6B~\citep{gpt-j} to generate pseudocode as intermediate FL expressions. We employ SymPy as the symbolic solver on top of Vicuna. \textcolor{black}{These two LMs are fine-tuned using their LM objective and PPO.} We build our own symbolic parser following the pseudocode structure discussed in Section~\ref{sec:formal-language-expression}.

We test \modelname\ with three different arithmetic problem datasets: SVAMP, GSM8K, and, MultiArith.
Details of the hyperparameters used in \modelname\ are described in  Appendix~\ref{app:Hyperparameters}.

\section{Experimental Results}
\label{sec:results}

We compare the performance of the LMs optimized using \modelname\ against multiple existing methods using LMs for solving arithmetic problems along with ablated versions of \modelname. In the following list of competitive methods, {\em Model-name} stands for Vicuna 13B and GPT-J: 
\begin{itemize}
    \item {\em Model-name} ({\bf ART}) implements the ART module proposed by \citet{paranjape2023art} that facilitates tool-usage along with {\em Model-name} as the base LM.
    \item {\em Model-name} ({\bf LMFT}) is the LM finetuned version of the base model optimized using Eq.~\eqref{eq:lm-finetune}.
    \item {\em Model-name} ({\bf k-shot}) is the base LM prompted with examples of FL generation samples and the symbolic solver; this provides the baseline ability of the LM without additional adapter-tuning or PPO; this approach is similar to PAL.
    \item Toolformer, as proposed by \citet{Toolformer}, is finetuned GPT-J with tool-usage.
    \textcolor{black}{
    \item {\em Model-name} ({\bf PAL}) implements the PAL module \cite{PAL} with {\em Model-name} as base LM.
    \item  {\em Model-name} ({\bf TRICE}) implements the TRICE module  \cite{qiao_making_2023} with {\em Model-name} as base LM.}
    
\end{itemize}
\begin{table}[!t]
\centering
\adjustbox{max width=.9\hsize}{\tabcolsep 2pt
\begin{tabular}{l||ccc} 
\multicolumn{1}{c||}{\bf Models} & \textbf{SVAMP} & \textbf{MultiArith} & \textbf{GSM8K} \\ \hline
  Vicuna 13B (\modelname) &  \(\bf 56.65\)   & \(\bf 59\)          &  \(\bf 35.2\)   \\ 

 Vicuna 13B (PAL) & \(53.7\) & \(49.4\) & \(27.5\) \\
 Vicuna 13B (ART) &  \(49.83\)   &  \(9.55\)       &   \(5.76\)   \\
 Vicuna 13B (TRICE) & - & \(18.83\) & - \\
 Vicuna 13B (LMFT) & \(42.5\)    &  \(48.3\)        & \(29.8\)    \\
  Vicuna 13B (4-shot) &  \(37.5\)   &     \(29\)    &  \(7.62\)  \\
 Vicuna 13B (1-shot) &  \(27\)   &      \(15.51\)    &  \(6.89\)  \\
 \hdashline
 GPT-J (\modelname) &  \(\bf 40.1\)   &   \(\bf 53.16\)  &  \(3.50\)   \\
 GPT-J (PAL) & \(22.33\) & \(6.79\) & \(\bf 8.25\)  \\
 GPT-J (LMFT) &  \(31.6\)   &  \(42.16\)      &   \(2.2\)   \\
 GPT-J (4-shot) &  \(9.45\)   &  \(2\)     &   \(0.61\)   \\ 
  GPT-J (1-shot) &  \(3\)   &  \(0.12\)     &   \(0\)   \\ 
 Toolformer &  \(29.4\)    &  -     &   -   \\
 GPT-J (ART) &  \(3.10\)   &  \(1.428\)       &   \(3.25\)   \\
 \hdashline
GPT-3.5 (standard 8-shot) & 64.8 & 34.0 & 15.1 \\
 \hline 
\end{tabular}
}
\caption{Accuracy (\%) of \modelname-optimized models against baselines. The performance of Toolformer on the SVAMP dataset is taken from their original paper \cite{Toolformer}; we could not compare \modelname\ against Toolformer on other datasets since the model weights/codebase are not publicly available. GPT-3.5 results are taken from \citep{zheng2023progressivehint}.}
\label{tab:main-result}
\end{table}

\begin{figure*}
    \centering
    \begin{tabular}{cp{.2\textwidth}}
    \includegraphics[width=.7\hsize, valign=t]{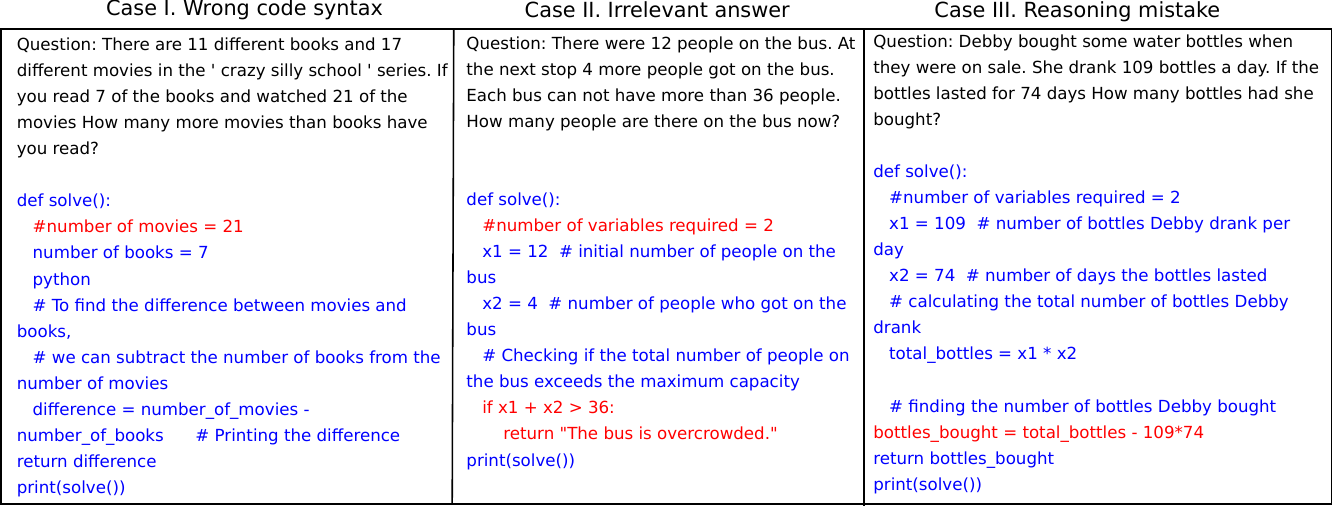}
    &
    \caption{Case-wise analysis of erroneous prediction by Vicuna 13B with \modelname\ from examples in SVAMP. Questions and generated Python codes are presented in black and blue, respectively, with the sources of error shown in red.}
    \label{fig:error}
    \end{tabular}
\end{figure*}

Table~\ref{tab:main-result} summarizes the performance of \modelname\ along with the baseline and ablation variants.
Vicuna 13B and GPT-J show strong improvements with \modelname\ optimization over their base versions with k-shot prompting. The performance gain is more staggering with GPT-J compared to Vicuna (in a zero-shot setting, GPT-J is not able to generate any pseudocode or Python code whatsoever). This aligns with the fact that the adapter finetuning with FL expression generation (Eq.~\ref{eq:lm-finetune}) essentially adds an instruction-following capability to the vanilla LM. We observe an interesting trend in \modelname\ vs LMFT across different datasets. With increasing order of difficulty (based on the performance of the base LM, MultiArith \(\rightarrow\) SVAMP \(\rightarrow\) GSM8K), the performance gain mostly increases. Intuitively, one can argue that the reasoning-specific rewards indeed help the model generalize better across harder tasks.

{\bf Reward ablation.} To investigate the role of the reward functions in the performance of \modelname, we remove each reward from \(R_1\) to \(R_4\) and run the PPO step.
Table~\ref{tab:Rl-ablation} summarizes the results of reward ablation. In case of Vicuna, the adverse effect of removing any of the reward function is evident; the order of importance can be deduced as \(R_4>R_3>R_2>R_1\). Given that Vicuna is an optimized code generation model, it is intuitive that the reward towards getting the correct answer (\(R_4\)) will be more influential than the reward to generate an executable code (\(R_1\)). Irrespective of the relative importance, these rewards play a crucial role in aligning the model towards better arithmetic reasoning. With GPT-J, however, the importance ordering is not that straightforward. On SVAMP, we observe a trend similar to Vicuna. On MultiArith, it looks like  the absence of \(R_1\) or \(R_2\) is beneficial, while results on GSM8K would suggest that removal of \(R_4\) actually helps. However, it should be noted that GPT-J in general performs in a near-random manner on GSM8K and any trend in such a low performance setup does not tell much of a story.

{\bf \modelname\ vs ART.} We discussed in Section \ref{sec:Rel} that ART presupposes a strong reasoning capability encoded within the LM itself (their reported results are primarily on 175B models). This assumption drastically fails with models having orders of magnitude smaller parameters like GPT-J. With GPT-J, ART exhibits near-random performance across all datasets. Appendix~\ref{app:ArtLimitations} displays GPT-J's inability to grasp reasoning strategies or tool applications. \textcolor{black}{With Vicuna 13B, the model performs well on SVAMP but fails to generalise on the other two datasets (MultiAirth and GSM8K)}. \modelname, on the other hand, is well suited for such small LMs as it provides a smooth interface between the LM and the symbolic solver via LoRA adapters. \textcolor{black}{\modelname\ is also able to handle datasets with varying difficulty.}

\begin{table}[h]
\small
        \centering
        \scalebox{0.9}{
        \begin{tabular}{l|c|ccc}
            Method & Model & SVAMP & MultiArith & GSM8K \\\hline
            \multirow{2}{*}{\modelname} & Vicuna 13B & 56.65 & 59 & 35.2\\
            & GPT-J & 40.1& 53.16 &3.50\\
            \hdashline
            \multirow{2}{*}{\modelname\(\setminus R_1\)} & Vicuna 13B & 51.0 & 52.3 & 34.95\\
            & GPT-J & 39.09& 57.66 &3.203\\
            \hdashline
             \multirow{2}{*}{\modelname\(\setminus R_2\)} & Vicuna 13B & 49.25 & 52.0 & 25.75\\
            & GPT-J & 39.78&54.16 &3.43\\
            \hdashline
            \multirow{2}{*}{\modelname\(\setminus R_3\)} & Vicuna 13B & 46.75 & 50.1 & 26.15\\
            & GPT-J & 35.69 & 47.5 & 3.02\\
            \hdashline
            \multirow{2}{*}{\modelname\(\setminus R_4\)} & Vicuna 13B & 31.08 & 45.8 & 22.2\\
            & GPT-J & 36.92 & 52 &3.81\\
            \hline
        \end{tabular}}
        \caption{Ablation study on \modelname\ with the reward functions. \modelname\(\setminus R_i\) denotes that the reward function \(R_i\) (see Section 4) is ablated.}
        \label{tab:Rl-ablation}
    \end{table}
    
 {\bf \modelname\ vs Toolformer.} 
  Contrary to ART, Toolformer is expected to achieve performance most similar to \modelname\ with GPT-J as the base model, given the fact that both methods use external tools while learning. The results on the SVAMP dataset verify the conjecture. However, even the simple pseudocode generation setup is much more powerful than learning to use a simple calculator like Toolformer. Furthermore, the PPO stage provides even stronger learning signals in \modelname, thereby showing a large performance improvement (\(+10.7\)) on the SVAMP dataset compared to Toolformer.
  
 {\bf Vicuna vs GPT-J.} The performance of all the GPT-J-based models compared to Vicuna reflects the difference in generalizability between vanilla LMs vs instruction-tuned ones. The base performance (k-shot) is near random across all the datasets. With the introduction of \modelname, we can observe a sharp rise in the performance on SVAMP (\textcolor{black}{+30.65}) and MultiArith (\textcolor{black}{+51.16}); however, that is not the case with GSM8K, where the performance gain, despite being positive, is meagre. Math-word problems in the latter dataset are qualitatively distinct compared to SVAMP and MultiArith. GSM8K is generally harder with more complex reasoning. The simplistic design of the pseudocode is not suited for such tasks as they tend to create a longer chain of statements that could have been expressed in Python more succinctly. Consequently, the required jump in generalized reasoning power is much higher for a pseudocode-generating GPT-J compared to a Python-generating Vicuna.

{\color{black}{\bf  {\modelname}\ vs PAL.} The experiment closest to PAL is {\em Model-name} (k-shot), which achieves lower performance when compared to PAL, but with \modelname\ (teaching how to use symbolic solver), it can surpass PAL performance. Therefore, PAL cannot learn how to use external tool usage to its full extent. GPT-J lacks the capability to generate Python code; simple pseudocode helps the small model in reasoning, thereby drastically increasing the performance compared to PAL. As mentioned above, GSM8K is harder; therefore, the pseudocode fails to express the complete reasoning steps.}

{\color{black}{\bf \modelname\ vs TRICE.} Vicuna 13B (TRICE) was trained from scratch with official code \cite{qiao_making_2023}. SVAMP and GSM8K were used as their training datasets; therefore, only MultiArith was used for comparison. \modelname{} shows a large improvement in the accuracy, pointing towards the advantage of converting problems to FL expression and then teaching the model to use the symbolic solver. }

{\bf Quality of NL to FL translation.} For GPT-J, the degree of syntactically incorrect pseudocodes generated are as follows: 2.3\% on SVAMP, 2.5\% on MultiArith, and 17.175\% on GSM8K. Note that GPT-J without \modelname\ is not able to generate a single, correct pseudocode even with few-shot example. With Vicuna, the base model produces 9\% syntactically incorrect python codes; with \modelname, this number becomes zero.

{\bf Error Cases with \modelname.} 
As we argued in Section~\ref{sec:formal-language-expression}, the generation template used in \modelname\ provides a better explanation as it synthesizes the flavor of CoT with the underlying PoT format. We analyze the prediction errors encountered by \modelname-optimized Vicuna 13B model. We randomly sampled 100 mispredictions from the SVAMP testbed and manually checked for possible sources of error. In Figure~\ref{fig:error}, three major error types are presented with one example each. In {\bf Case~I}, the model is struggling to generate correct syntax; it misses the number of variables needed as a comment after generating the start-of-comment token \#. Instead, it directly initializes the variables; as a result, the generated code throws a compilation error with the usage of undefined variables ({\it number\_of\_movies} in this example). This is very much a sign of the LM overlooking details of the instructions and following its pretraining character. Next, we have {\bf Case~II}, with the model generating irrelevant answers to the question. In this example, the model generates the answer to whether or not the bus is overcrowded, while the question is to count the number of people in the bus. The model fell prey to the misdirect presented in the question stating the maximum capacity of the bus and ignored the later part. While this failure is somewhat blunt in the manifestation in this example, even huge LMs like ChatGPT are not robust against such adversarial misdirects while reasoning~\citep{borji2023categorical}. Finally, in {\bf Case~III}, we have straightforward reasoning errors.  
\begin{figure}[!t]
  \begin{center}
    \includegraphics[width=\hsize]{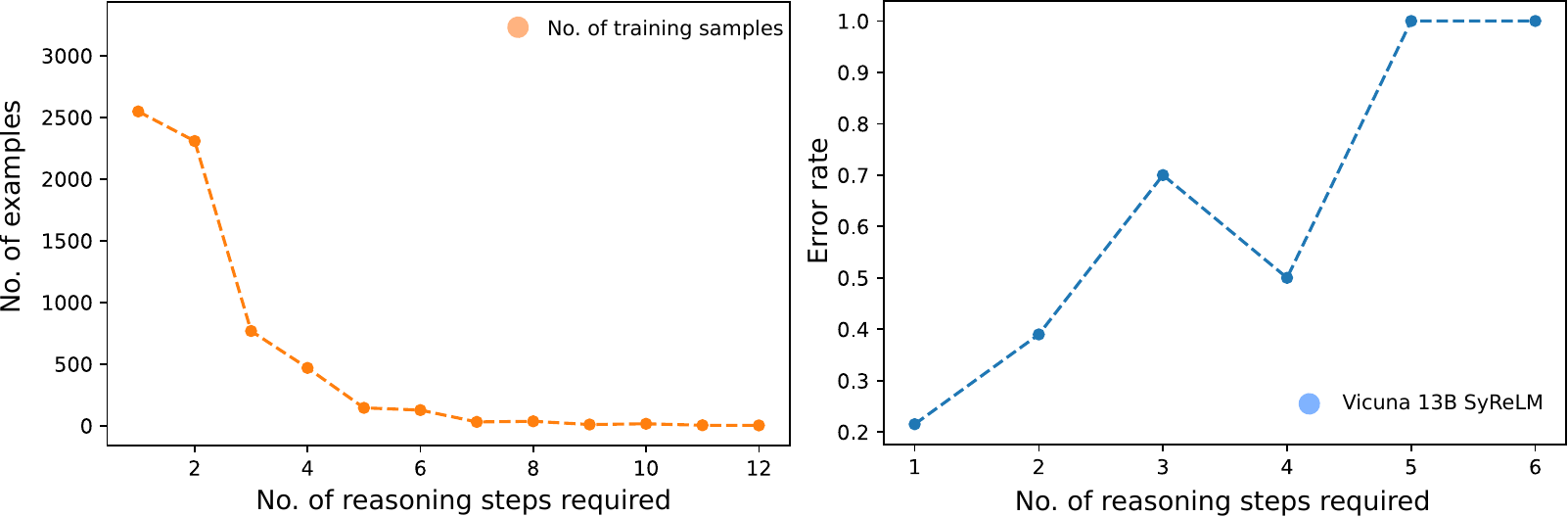}
  \end{center}
  \caption{({\bf Left}) Distribution of training data examples across the number of reasoning steps required (number of lines in the Python code). ({\bf Right}) The prediction error rate of Vicuna 13B with \modelname\ vs the number of reasoning steps required to solve the problem.} \label{fig:reasoning-step-example}
\end{figure}
The model essentially lost the reasoning chain in this example; it correctly identified how to calculate the total number of bottles. Yet, it could not identify that the total number of bottles bought is the same as the total number of bottles drunk. Additional analyses with examples generated by GPT-J with \modelname\ are presented in Appendix~\ref{app:SVAMPErrorAnalysis}. Similar error analyses on examples from GSM8K and MultiArith are provided in Appendix~\ref{app:GSM8KandMultiArithErrors}.

{\bf Generalizability across Reasoning Steps.} Figure~\ref{fig:reasoning-step-example} (left panel) shows the distribution of the reasoning steps required in the training data (count of non-comment lines in the generated Python script). Additionally, we use a subset of the SVAMP dataset to annotate the Python scripts using similar methods as described in Section~\ref{sec:experiments} to estimate the number of reasoning steps required to solve the problems.  Figure~\ref{fig:reasoning-step-example} (right panel) shows the error rate of Vicuna 13B with \modelname\ on problems requiring different reasoning steps. While the model generally shows an increasing error rate with an increased number of reasoning requirements, it is still able to generalize to longer reasoning chains that are rarely represented in the training data. This supports our hypothesis of increased generalizability with \modelname.

\section{Conclusion}
\label{sec:End}
Notwithstanding great recent enthusiasm about LLMs as all-purpose problem solvers, practitioners appreciate that LLMs work best when limited in their role to act as a glue between tools specialized to non-linguistic tasks like logic, arithmetic, or structured information retrieval. In response, LLMs are steadily getting better at invoking tools. Here, through the design of a new system, \modelname, we explore a synergy between symbolic and numeric reasoning that has been established in middle-school pedagogy for a while, but not yet commonplace with LLMs. \modelname{} is based on a frozen LM of modest size, coupled with a low-rank adapter for fine-tuning, keeping the setup within the computational capacity of most research groups. By prompting \modelname{} to build a bridge between chain-of-thoughts and formal programs, we show that even frugal LMs can be effective at solving complex, multi-step arithmetic word problems, a capability thought to be emergent only in LLMs of massive sizes trained on enormous amounts of data.

 {\bf Limitations.} Our work primarily focuses on frugal LMs, which, as our experiments and existing literature suggest, have limited language generation and abstract reasoning capability. While our proposed method \modelname\ improves greatly upon their base performance, some implicit shortcomings of these LMs manifest across the optimized versions as well. The LoRA finetuning process along with PPO requires additional GPU-hours compared to the fully zero-shot generation of the API-based large LMs.
\section*{Acknowledgments}
The authors acknowledge the financial support of DYSL-AI.
\bibliography{aaai24}

\cleardoublepage
\appendix
\newtcolorbox{promptbox}[1]{
  colback=gray!8,
  colframe=black!70,
  title={\raggedright #1},
  coltitle=white,
  enhanced,
  arc=.5mm,
  boxsep=0mm,
  left=1mm,
  fontupper=\small,
  hbox, 
  capture=minipage,
  width=0.5\textwidth 
}

\begin{center}
{\Large\bfseries\ztitle} \\
{\large\bfseries (Appendix)}
\end{center}


\section{Details of training dataset}
\label{app:TrainingDataDetails}

All pseudocode examples use a few patterns:
\begin{enumerate}
\item The {\bf [find]} operation is used to extract numeric values in the problem statement.
\item \label{step:reason} The reasoning phase invokes operations like {\bf [multiply]}, {\bf [divide]}, {\bf [add]}, {\bf [subtract]} on input and derived values.
\item \label{step:compute} Computation steps include a comment in natural language explaining the intent of the computation.
\item The last step returns the target value as answer.
\end{enumerate}
Steps \ref{step:reason} and \ref{step:compute} can occur multiple times.
There are a total of 9 major operations in the training dataset; some are rarely used.
\begin{center}
\begin{tabular}{|c|r|} \hline
Operator & Frequency \\ \hline
multiply & 3950 \\
divide & 2931 \\
add & 3171 \\
subtract & 3125 \\
lcm & 104 \\
gcd & 90 \\
round & 74 \\
floor & 12 \\
mod & 9 \\ \hline
\end{tabular}
\end{center}

Following are a few examples of arithmetic problems and their corresponding pseudocode solutions used in the training dataset.(- denotes new line)
\begin{promptbox}
{Question: Xiaogang scored 18 goals, and Xiaoqiang scored twice as many goals as Xiaogang. How many goals did they score together?}

\begin{verbatim}
Pseudocode:
- var1 = [find](number of goals 
scored by Xiaogang) #18
- var2 = [multiply](var1, 2) # 18 * 2 = 36
- var3 = [add](var1, var2) # 18 + 36 = 54
- [return](var3) # 54    
\end{verbatim}
\end{promptbox}

\begin{promptbox}{Question: The power supply bureau originally had bundles of wires that were 14.85 meters long. The master worker used up 0.75 meters for the first time, and 1.25 meters for the second time. How many meters of wires are left at this time?}
\begin{verbatim}
Pseudocode:
- var1 = [find](length of original 
bundle of wires) #14.85
- var2 = [find](length used for 
the first time) #0.75
- var3 = [find](length used for 
the second time) #1.25
- var4 = [add](var2, var3) 
# 0.75 + 1.25 = 2
- var5 = [subtract](var1, var4) 
# 14.85 - 2 = 12.85
- [return](var5) # 12.85
\end{verbatim}
\end{promptbox}

\begin{promptbox}{Question: In the pattern rope skipping activity, 80 people skipped the long rope, and 40 people skipped the short rope twice as many as the long rope. How many people jump rope?}
\begin{verbatim}
Pseudocode:
- var1 = [find](number of people 
skipping the long rope) #80
- var2 = [find](number of people skipping the 
short rope in relation to the long rope) #2
- var3 = [divide](var1, var2) # 80 / 2 = 40
- var4 = [multiply](var3, var2) # 40 * 2 = 80
- var5 = [add](var3, var4) # 40 + 80 = 120
- [return](var5) # 120
\end{verbatim}
\end{promptbox}

\begin{promptbox}{Question: A road is 570 meters long, has been repaired for 8 days, and there are 250 meters left. How many meters are repaired per day on average?}
\begin{verbatim}
PseudoCode:
- var1 = [find](length of road) #570
- var2 = [find](total days repaired) #8
- var3 = [find](distance left to repair) #250
- var4 = [subtract](var1, var3) 
# 570 - 250 = 320
- var5 = [divide](var4, var2) 
# 320 / 8 = 40
- [return](var5) # 40
\end{verbatim}
\end{promptbox}

\begin{figure}[h]
    \centering
    \includegraphics[width=\linewidth]{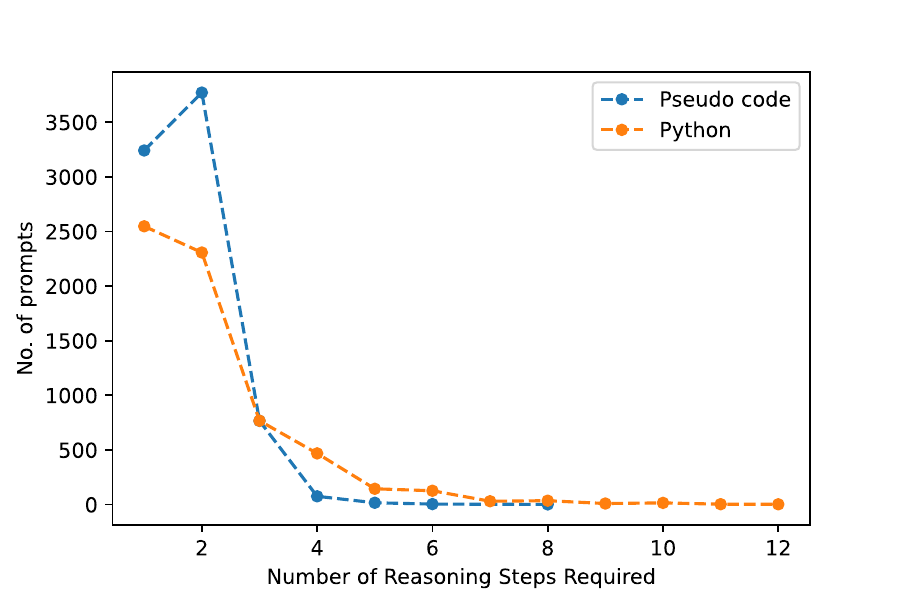}
    \caption{Distribution of training data problems against the number of reasoning steps needed to solve them (number of program statements).}
    \label{fig:step-distribution}
\end{figure}

In Figure~\ref{fig:step-distribution}, we plot the distribution of the number of reasoning steps required in Python and Pseudocode programs in the training dataset used. 

\section{Hyperparameters and training policies}
\label{app:Hyperparameters}

All the models were implemented using Huggingface\footnote{https://huggingface.co/} with PyTorch. To implement the PPO algorithm, we make use of the TRL library~\footnote{https://github.com/lvwerra/trl}.
We use a single NVIDIA A100 (80 GB) GPU for all the training and testing purposes.

\paragraph{Vicuna 13B} For the LoRA finetuning of Vicuna 13B model using LM objective, we use the following hyperparameter values: Micro batch size= 8, Batch size = 64, Gradient accumulation steps = 8, No. of epochs trained = 5, Optimizer used = Adam, Learning rate = 2e-5 
Maximum input length = 2048, LoRA $r$ = 4,
LoRA $\alpha$ = 8, LoRA dropout = 0.05.

For PPO on Vicuna 13B with LoRA adapters, we use the following hyperparameter values: Learning Rate = 1.41e-6, Init\_kl\_coeff = 0.03 (Initial KL penalty coefficient $\beta$ used for adaptive and linear control), $\gamma$ = 0.99, Batch size = 64, Mini batch size = 2, Steps = 20000, Adap\_kl\_ctrl = True, Target = 6 (Target KL value for adaptive KL control), Horizon = 10000 (Horizon for adaptive KL control), $\lambda$ = 0.95 (Lambda parameter for advantage calculation), Cliprange = 0.2 (Range for clipping in PPO policy gradient loss), Cliprange value = 0.2 (Range for clipping values in loss calculation), epochs = 3, Target\_kl = 0.05.

In the inference time, the following generation parameters were used: top\_k = 1, top\_p = 1, max\_new\_tokens = 512, Temperature = 1, Do\_sample = False.

\paragraph{GPT-J} For the LoRA finetuning of GPT-J model using LM objective, we use the following hyperparameter values: Batch size = 8, No. of epochs trained = 10, Optimizer used = Adam, Learning rate = 5e-6, 
Maximum input length = 256, LoRA $r$ = 16,
LoRA $\alpha$ = 32, LoRA dropout = 0.05.

For PPO on GPT-J with LoRA adapters, we use the following hyperparameter values: Learning Rate = 1.41e-6, Init\_kl\_coeff = 0.03 (Initial KL penalty coefficient $\beta$ used for adaptive and linear control), $\gamma$ = 0.99, Batch size = 128, Mini batch size = 4, Steps = 20000, Adap\_kl\_ctrl = True, Target = 6 (Target KL value for adaptive KL control), Horizon = 10000 (Horizon for adaptive KL control), $\lambda$ = 0.95 (Lambda parameter for advantage calculation), Cliprange = 0.2 (Range for clipping in PPO policy gradient loss), Cliprange value = 0.2 (Range for clipping values in loss calculation), epochs = 4, Target\_kl = 0.1.

At inference time, the following generation parameters were used: top\_k = 4, top\_p = 0.7, max\_new\_tokens = 120, Temperature = 1, Do\_sample = False, Num\_beams = 4.
\section{Limitations of ART}
\label{app:ArtLimitations}

In a few-shot setting,  Vicuna and GPT-J generate the output in a format parsable by ART but fails to generate correct in GSM8K dataset. We provide a few examples of the outputs generated by both these models on problems from the SVAMP dataset as follows.

\begin{promptbox}{Question: A pet store had 13 siamese cats and 5 house cats. During a sale they sold 10 cats. How many cats do they have left?}

\paragraph{Vicuna:}
\begin{verbatim}
Q1:Write a python code to find the number 
of cats the pet store has left.
#1:
total_cats = 13 + 5
cats_sold = 10
left_cats = total_cats - cats_sold
print(left_cats)
Q2: Execute the python code 
and get the answer.
#2: 18
Q3: [EOQ]
Ans: 18
\end{verbatim}

\paragraph{GPT-J}:
\begin{verbatim}
Q1:[generate python code] write down the 
arithmetic or algebra equations as 
python code, storing the answer as 'ans'
#1:
number_of_cats_left = 13 - 10
ans = number_of_cats_left
print(ans)
Q2: [code execute] Execute the python code 
in #1 and get the value of "ans"
#2: 5
Q3: [EOQ]
Ans: 5
\end{verbatim}
\end{promptbox}

\begin{promptbox}{Question: The school cafeteria ordered 42 red apples and 7 green apples for students lunches. But, if only 9 students wanted fruit, how many extra did the cafeteria end up with?}
\paragraph{Vicuna:}
\begin{verbatim}
Q1:Write a python code to find the number of 
extra apples the cafeteria has.
#1:
total_apples = 42 + 7
extra_apples = total_apples - 9
ans = extra_apples
print(ans)
Q2: Execute the python code and 
get the value of "ans"
#2: 33
Q3: [EOQ]
Ans: 33
\end{verbatim}

\paragraph{GPT-J:}
\begin{verbatim}
Q1:[generate python code] write down the 
arithmetic or algebra equations as 
python code, storing the answer as 'ans'
#1:
num_of_apples_ordered = 42
num_of_apples_left = 7
num_of_apples_ordered - num_of_apples_left=9
ans = num_of_apples_left
print(ans)
Q2: [code execute] Execute the python code 
in #1 and get the value of "ans"
#2: 7
Q3: [EOQ]
Ans: 7
\end{verbatim}
\end{promptbox}

\begin{promptbox}{Question: Carol and her mom were picking carrots from their garden. Carol picked 29 and her mother picked 16. If only 38 of the carrots were good, how many bad carrots did they have?}
\paragraph{Vicuna:}
\begin{verbatim}
Q1:[generate python code] write down the 
arithmetic or algebra equations as 
python code, storing the answer as 'ans'
#1:
carrots_picked_by_carol = 29
carrots_picked_by_mom = 16
good_carrots = 38
bad_carrots = total_carrots-good_carrots
ans = bad_carrots
print(ans)
Q2: [code execute] Execute the python code 
in #1 and get the value of "ans"
#2:
bad\_carrots = 38
print(bad\_carrots)
Q3: [EOQ]
Ans: 38
\end{verbatim}

\paragraph{GPT-J:}
\begin{verbatim}
Q1:[generate python code] write down the 
arithmetic or algebra equations as 
python code, storing the answer as 'ans'
#1:
good_carrots = 38
bad_carrots = 29
total_carrots = good_carrots + bad_carrots
ans = total_carrots - bad_carrots
print(ans)
Q2: [code execute] Execute the python code 
in #1 and get the value of "ans"
#2: 38
Q3: [EOQ]
Ans: 38 
\end{verbatim}
\end{promptbox}

\section{Anecdotes and error analyses: SVAMP}
\label{app:SVAMPErrorAnalysis}

\begin{figure*}[t]
    \centering
    \includegraphics[width=0.9\textwidth]{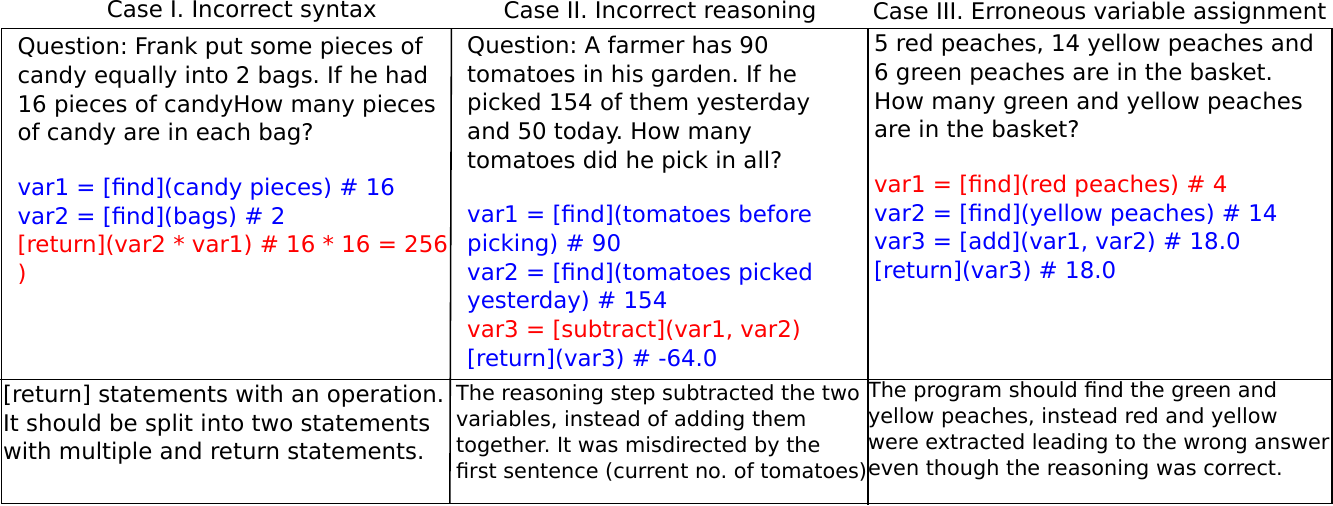}
    \caption{Analysis of errors done by \modelname-optimized GPT-J on SVAMP dataset. We provide one example for each type of error (question in black, generated pseudocode in blue with erroneous steps marked in red). Short explanations of the error specifics are also given.}
    \label{fig:error-gpt-svamp}
\end{figure*}

We analyze errors generated by \modelname-optimized GPT-J on the SVAMP dataset, similar to Figure~\ref{fig:error}. We could identify three major sources of errors that are very similar to Vicuna 13B:
\begin{enumerate*}[(i)]
\item {\bf syntax error},
\item {\bf incorrect reasoning}, and
\item {\bf erroneous variable assignment} by {\bf [find]} statements; 
\end{enumerate*}
examples are provided in Figure~\ref{fig:error-gpt-svamp}.

\begin{figure*}[t]
    \centering
    \includegraphics[width=0.9\textwidth]{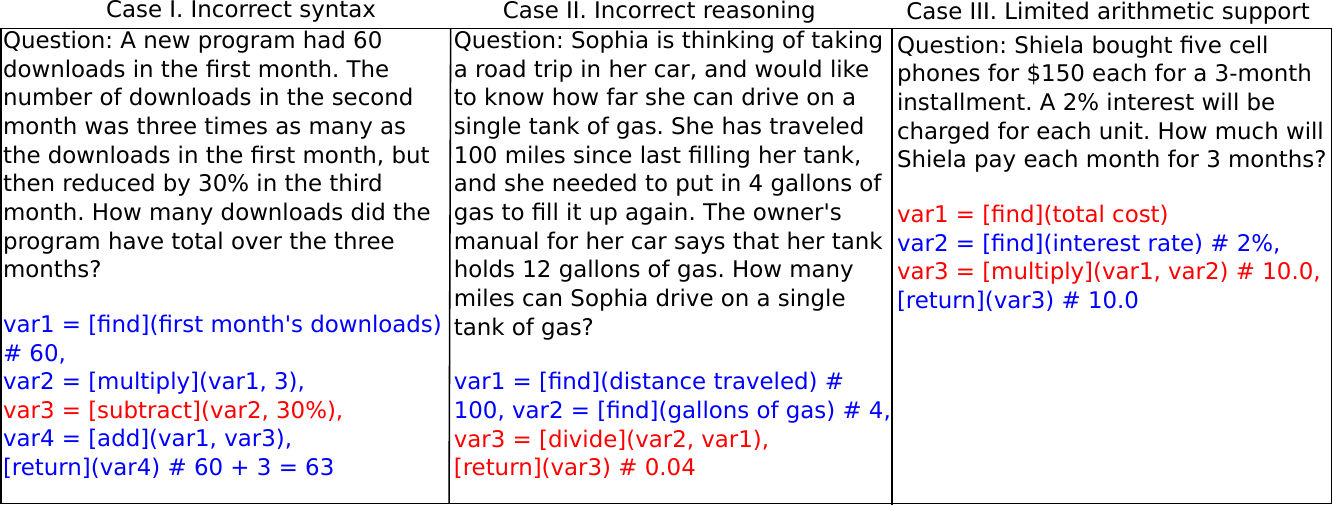}
    \caption{Error analysis of \modelname-optimized GPT-J on GSM8K.}
    \label{fig:error-gsm8k}
\end{figure*}

\begin{figure*}[t]
    \centering
    \includegraphics[width=0.6\textwidth]{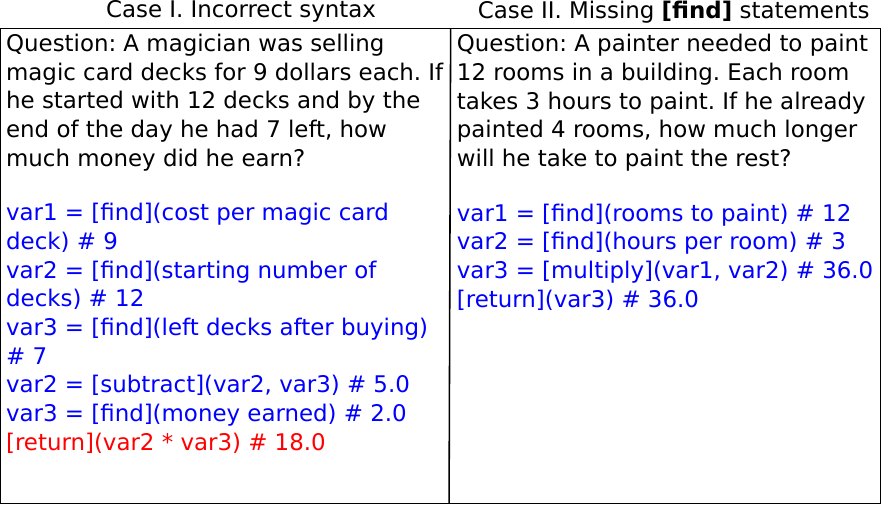}
    \caption{Error analyses with \modelname-optimized GPT-J on MultiArith. We show the two most common types of errors here. In Case II, the model could not identify the value of rooms painted in the {\bf [find]} statements.}
    \label{fig:my_label}
\end{figure*}

\section{Anecdotes and error analysis: GSM8K and MultiArith}
\label{app:GSM8KandMultiArithErrors}

\begin{figure*}[h]
    \centering
    \includegraphics[width=\textwidth]{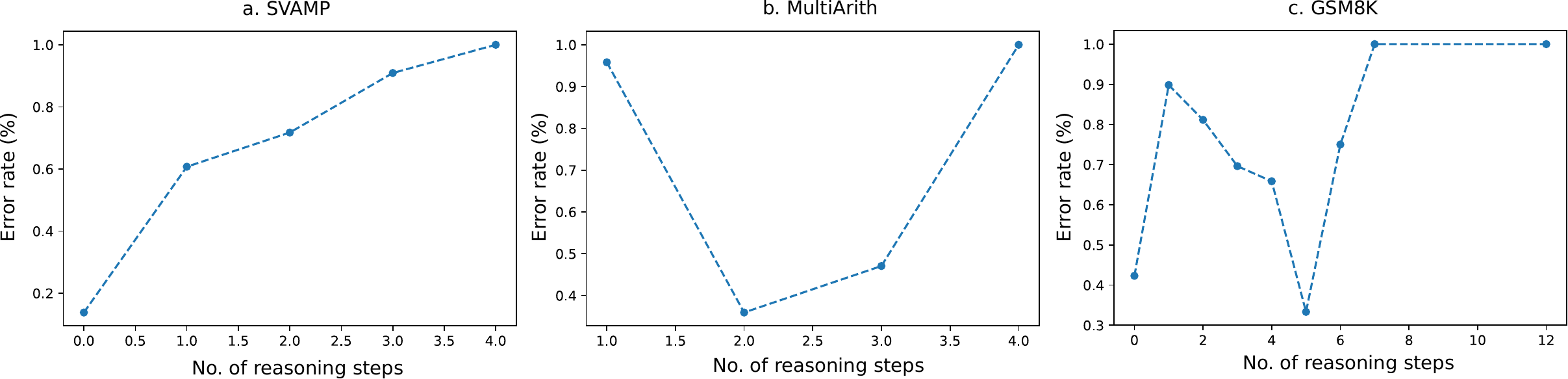}
    \caption{Error rates with \modelname-optimized GPT-J at problems requiring different numbers of reasoning steps across all three datasets.}
    \label{fig:error-rate-all}
\end{figure*}
In Figure~\ref{fig:error-gsm8k}, we show three examples of errors originating from \modelname-optimized GPT-J on the GSM8K dataset. Most errors on GSM8k are due to its long and complex questions; we have the largest number of programs with incorrect syntax in this dataset ---
15\% in GSM8K compared to 2.6\% in MultiArith and 5.17\% in SVAMP. The average length of a prompt is the largest for GSM8k, making it harder for smaller LLMs such as GPT-J to parse, as shown in case~II of Figure~\ref{fig:error-gsm8k} (average number of tokens in the prompt: 33.82 tokens in MultiArith, 30.05 tokens in SVAMP, 47.01 tokens in GSM8K). GSM8K also contains values with percentages and interest, which are complex and cannot be handed due to the limitation of the pseudocode (Case~I and III in Figure~\ref{fig:error-gsm8k}).

In case of the MultiArith dataset, the error patterns are similar to SVAMP. We show two of the most common types of errors in Figure~\ref{fig:my_label}.
In Figure~\ref{fig:error-rate-all}, we show how the error rate changes for \modelname-optimized GPT-J across different numbers of reasoning step requirements.

\end{document}